\documentclass[10pt,twocolumn,letterpaper]{article}

\usepackage[pagenumbers]{cvpr} 

\definecolor{cvprblue}{rgb}{0.21,0.49,0.74}
\usepackage[pagebackref,breaklinks,colorlinks,allcolors=cvprblue]{hyperref}
\usepackage[accsupp]{axessibility}  
\usepackage{amsmath}
\usepackage{booktabs}
\usepackage{amssymb}
\usepackage{multirow}
\usepackage{standalone}
\usepackage{makecell}
\usepackage[table]{xcolor}
\usepackage{algorithm}
\usepackage{algorithmic}
\usepackage{graphicx}
\usepackage{listings}
\usepackage{multicol}
\definecolor{lightgray}{gray}{0.95}
\definecolor{midgray}{gray}{0.88}

\usepackage[most]{tcolorbox}
\newtcolorbox{promptbox}[2][]{
  enhanced,
  title={#2},
  colframe=black!75,    
  colback=blue!5!white, 
  coltitle=white,
  fonttitle=\bfseries,
  attach boxed title to top left={xshift=5mm, yshift*=-3mm},
  boxed title style={colframe=black!75, colback=black!75},
  top=10pt, bottom=10pt, left=10pt, right=10pt,
  arc=3mm,
  #1
}

\usepackage[absolute,overlay]{textpos} 
\def\paperID{18629} 
\def\confName{CVPR}
\def\confYear{2026}

\title{Training High-Level Schedulers with Execution-Feedback Reinforcement Learning for Long-Horizon GUI Automation}

\author{
    Zehao Deng\textsuperscript{1,2}\thanks{Work done during Zehao Deng's visit at Shanghai Jiao Tong University. $\dagger$ Corresponding author.}\hspace{0.4em}, 
    Tianjie Ju\textsuperscript{2}, 
    Zheng Wu\textsuperscript{2}, 
    Zhuosheng Zhang\textsuperscript{2$\dagger$}, 
    Gongshen Liu\textsuperscript{2}
    \vspace{2mm} \\ 
    \textsuperscript{1}School of Computer Science and Technology, Soochow University \\
    \textsuperscript{2}School of Computer Science, Shanghai Jiao Tong University \\
    {\ttfamily\small 2327406010@stu.suda.edu.cn, \{jometeorie,wzh815918208,zhangzs,lgshen\}@sjtu.edu.cn}
}

\begin{document}

\maketitle


\begin{abstract}
The rapid development of large vision-language model (VLM) has greatly promoted the research of GUI agent. However, GUI agents still face significant challenges in handling long-horizon tasks. First, single-agent models struggle to balance high-level capabilities and low-level execution capability, facing prevalent issues of responsibility coupling and capability conflicts. Second, agents lack awareness of the task state, leading to progress loss in long-horizon tasks. 
To address these challenges, we propose a staged execution-feedback reinforcement learning algorithm. Unlike training a unified policy model, we focus on training high-level scheduling models. Specifically, we propose and train two agents: a Coordinator, responsible for the strategic planning and task decomposition; and a State Tracker, responsible for context compression and information management to maintain the task's state and coherence. Based on this, we built the Coordinator-Executor-State Tracker (CES) multi-agent framework, which can be integrated with any low-level Executor model, assisting the Executor in solving long-horizon tasks through task scheduling and state management.
Experiments on long-horizon task benchmarks demonstrate that CES significantly enhances the system's planning and state management capabilities. Furthermore, analysis confirms that our trained high-level scheduling module is a generalizable, plug-and-play module that significantly enhances the long-horizon capabilities of various Executors. Code is available at \href{https://github.com/hehehahi4/CES}{https://github.com/hehehahi4/CES}.
\end{abstract}

\begin{figure}[t!]
    \centering
    \includegraphics[width=\columnwidth]{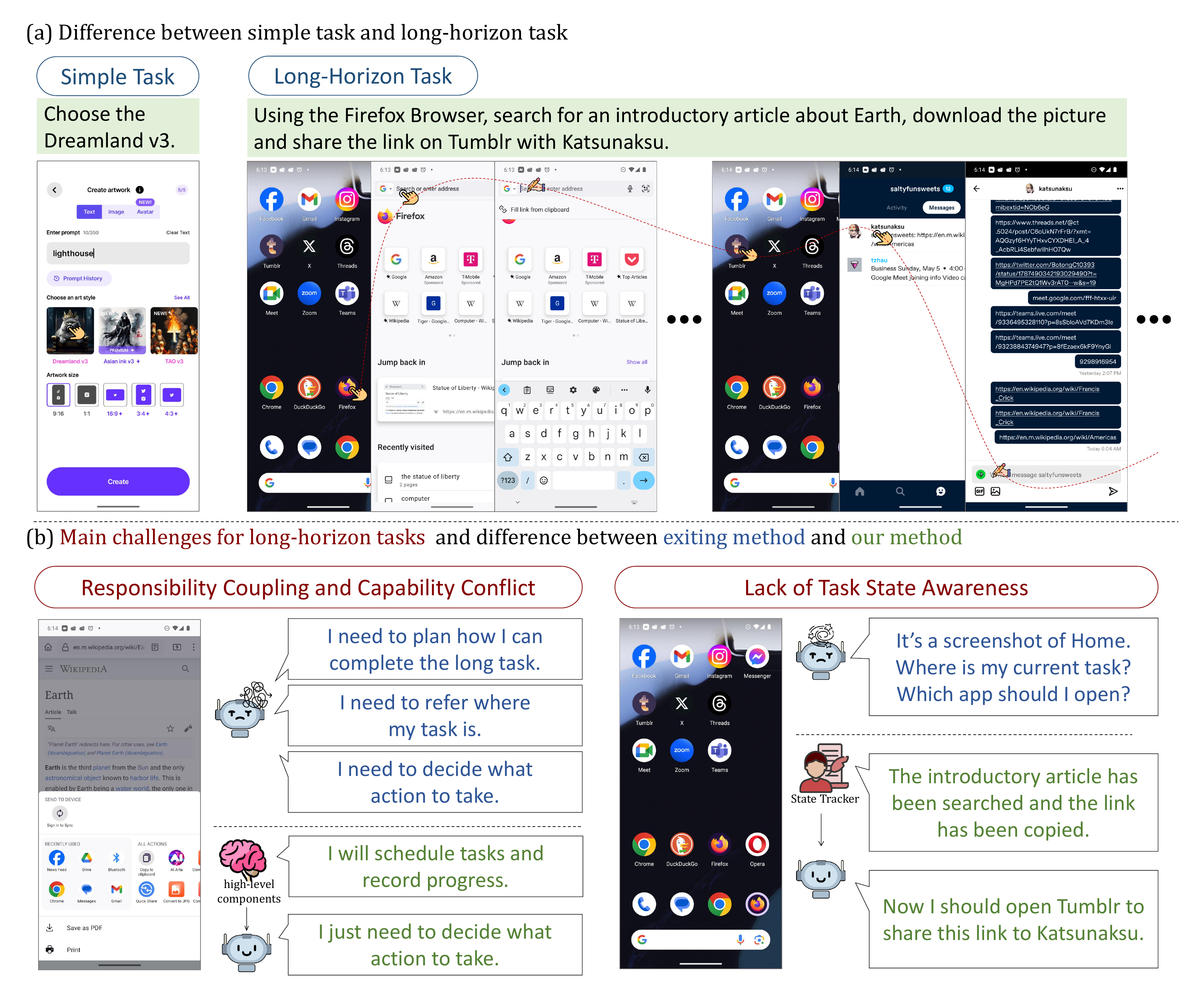} 
    \caption{(a) Difference between simple tasks and long-horizon tasks. A simple task only involves a single action on one screen driven by atomic instruction, whereas a long-horizon task requires a complex trajectory driven by ambiguous, high-level user instruction. (b) Comparison of how existing methods and our method address long-horizon challenges. Left (Responsibility Coupling and Capability Conflict): A single agent is overloaded by coupling high-level capability and low-level execution. Our method resolves this by decoupling these roles into high-level and low-level components. Right (Lack of Task State Awareness): A single agent loses context on ambiguous screens like the Home screen. Our State Tracker provides high-semantic memory, enabling correct, context-aware decisions.}
    \label{fig:intro}
\end{figure}

\section{Introduction}
Graphical User Interface (GUI) agents play a crucial role in automating complex tasks \cite{agashe2025agent, cheng2025oskairos, li2025coloragent, rawles2025androidworld, sun2025osgenesis, wu2025verios, xu2025aguvis, tang2025surveymllmbasedguiagents}. Traditional training paradigm for GUI agents primarily relies on Supervised Fine-Tuning (SFT), where the model learns the mapping from environmental states to actions through imitation learning \cite{wu2024osatlas, qin2025uitars, zhang2025agentcpmgui}. However, SFT heavily depends on large-scale, costly, and meticulously annotated high-quality trajectory data, and often exhibits poor generalization capability in unseen environments \cite{lai2025computerrl, lu2025uir1, luo2025guir1, wei2025webagentr1}.

Recent studies \cite{li2025chainofagents, luo2025guir1, wu2025backtrackagent, zhang2025rlvmr} have largely focused on using rule-based Reinforcement Learning (RL) to train agents. They utilize verifiable reward functions, supplanting the need for expensive manual annotations and human feedback. RL has been demonstrated as an efficient alternative to SFT in GUI tasks, as it requires only a relatively small number of high-quality training data to achieve significant performance gains \cite{luo2025guir1}. While existing methods have achieved commendable results in simple tasks \cite{liu2025infiguir1, lu2025uir1, luo2025guir1}, they still face two fundamental problems when confronted with long-horizon tasks, illustrated in \Cref{fig:intro}:

(i) \textbf{Responsibility Coupling and Capability Conflict in Single-Agent Architectures.} Current mainstream end-to-end model attempts to couple heterogeneous capabilities, such as long-term task planning, multi-step reasoning, GUI element grounding, and precise action execution, within a unified policy network. This design in optimization presents fundamental difficulties: a model with finite parameters struggles to simultaneously master high-level abilities, such as decomposing complex instructions, tracking task progress, alongside low-level abilities like grounding and execution. As task complexity increases, this coupling may lead to a catastrophic collapse of the model's diverse capabilities.

(ii) \textbf{Lack of Task State Awareness.} In long-horizon tasks, an accurate awareness of the current progress is crucial for making correct decisions. Most methods rely on historical action sequences (e.g., \texttt{Click(x, y)}), but these low-level actions provide almost no task state information or semantic context. Therefore, the agent has to primarily rely on the visual information from screenshots to infer its progress. However, screenshots are an insufficient and unreliable representation of task progress. This limitation makes it difficult for the agent to determine its current position in a long task, leading to errors and progress loss.

To address the challenges above, we propose a staged execution-feedback RL algorithm, where the insight is to resolve responsibility coupling and conflicting optimization objectives.
To support this algorithmic paradigm, we built a Coordinator-Executor-State Tracker (CES) framework, which structurally decouples the complex automation process into three specialized agents.
Unlike training a unified and overloaded policy model, our algorithm is designed to optimize specific high-level scheduling models. Specifically, we treat the low-level Executor as a frozen, plug-and-play component to provide verifiable feedback, and focus our training exclusively on the Coordinator and State Tracker. Through our staged optimization strategy, the Coordinator is trained to handle strategic planning and task decomposition, effectively decoupling high-level reasoning from low-level execution; meanwhile, the State Tracker is trained to act as dynamic memory, directly solving the lack of task state awareness by maintaining a high-semantic state summary in natural language.

In summary, our main contributions are as follows:
\begin{itemize}
    \item We build CES multi-agent framework, featuring general-purpose, plug-and-play high-level components (Coordinator and State Tracker) that can integrate with various Executors and enhance their abilities.
    \item We introduce a State Tracker, whose core task is dynamic context compression and state summarization, effectively resolving the state unawareness problem and maintaining the agent's logical coherence in long-horizon tasks.
    \item We propose a staged execution-feedback RL strategy. The core of this algorithm is to decouple high-level capabilities from low-level execution: it freezes a pre-trained Executor and uses the reward signals it generates to exclusively train the high-level Coordinator and State Tracker.
    \item Extensive experiments demonstrate that our method significantly enhances the long-horizon scheduling and state management capabilities of various Executor models and surpasses existing baselines.
\end{itemize}

\section{Related Work}

\subsection{GUI Agent}
Traditional mobile intelligent assistants primarily relied on rule-based or intent-driven API calls or structured text representations \cite{fan2025appcopilot, li2025survey, zhang2024large}. However, these methods are often platform-specific \cite{wang2025uitars2}, difficult to generalize \cite{luo2025guir1}. With the development of VLMs, research has shifted towards pure-vision settings \cite{li2025survey, yang2025zerogui, xu2025aguvis}. In this pure-vision paradigm, the agent receives only screenshots and task descriptions as input, generating coordinate-based actions directly in pixel space \cite{cheng2025oskairos, gu2025uivenus, hong2024cogagent, zhang2024large}.

Early methods \cite{cheng2025oskairos, rawles2025androidworld, sun2025osgenesis, wu2024osatlas} commonly relied on the SFT paradigm, training models via imitation learning. However, the SFT paradigm has two major limitations: dependency on large-scale annotated data and poor generalization \cite{luo2025guir1, shi2025mobileguirl, sun2025seagent, wei2025webagentr1, zhang2025rlvmr}.

\subsection{RL-Based GUI Agent}

To enable agents to learn more strategic decision-making, recent research has gradually shifted to RL, which is a key paradigm for enhancing the planning and decision-making capabilities of GUI agents in difficult task scenarios \cite{li2025survey, lu2025uir1, shi2025mobileguirl}. RL involves fine-tuning pretrained VLMs using relatively small-scale, high-quality interaction trajectories, thereby enhancing their specialized reasoning and decision-making skills without compromising their original capabilities \cite{li2025coloragent, xu2025mobilerl, ye2025mobileagentv3}. Inspired by the success of DeepSeek-R1 \cite{deepseek-ai2025deepseekr1}, the Group Relative Policy Optimization (GRPO) \cite{shao2024deepseekmath} algorithm has been widely adopted in the GUI agent domain. GUI-R1 \cite{luo2025guir1} is a representative model that applies R1-style reinforcement learning to enhance the capabilities of VLMs in high-level, real-world GUI tasks.

However, these RL-based single-agent methods \cite{liu2025infiguir1, lu2025uir1, luo2025guir1, shi2025mobileguirl, sun2025seagent, tang2025gui, lu2025ui}, while improving the decision-making algorithm, fail to address a fundamental problem: the capability overload of a single policy network. We argue that whether using SFT or RL, coupling heterogeneous abilities like high-level strategic planning with low-level visual perception and precise execution into the same model inevitably leads to optimization conflicts and role confusion. Therefore, we innovate through architecture, thoroughly decoupling different responsibilities to enhance all facets of the agent's capabilities via specialized division of labor and collaboration.

\subsection{Multi-agent Framework}
Multi-agent systems represent a key trend in the development of LLM-driven agents \cite{zhang2025mobiagent}, overcoming the inherent limitations of single agents in handling complex, dynamic, and long-horizon tasks through specialized division of labor and intelligent collaboration \cite{li2025mobileuse, wu2025hiagent}. For example, Mobile-Agent-v3 \cite{ye2025mobileagentv3} coordinates four agents to achieve robust, adaptive task automation with knowledge evolution capabilities. MobiAgent \cite{zhang2025mobiagent} adopts a Planner-Decider-Grounder multi-agent architecture, combined with an acceleration framework to improve execution accuracy.

However, existing works \cite{guo2025atomictocompositional, li2025appagent, ye2025mobileagentv3} often use general VLMs, employing prompt engineering to make models play different roles. This collaboration lacks deep optimization for each role, making it difficult to achieve optimal efficiency and specialized capability. Furthermore, existing research often overlooks the importance of task state management and information compression in long-horizon tasks, merely using historical actions \cite{lu2025swirl, shi2025mobileguirl} or passively storing large amounts of redundant information \cite{gao2025chainofmemory, kong2025mapagent}. Our CES framework, by introducing a dedicated State Tracker and an execution-feedback RL strategy, provides specialized and efficient optimization paths for each multi-agent role.

\section{Preliminary Experiment}
In the Introduction, we posited a key hypothesis: GUI agents fail in long-horizon tasks due to a lack of task state awareness, arguing that screenshots are an insufficient and unreliable representation of task progress.

To empirically validate this specific hypothesis, we designed a temporal reasoning experiment. This experiment is built upon a key logic: if a screenshot were a sufficient representation of state, an agent should be able to reliably determine its relative position in a task's timeline. Our experiment directly tests this ability by tasking three powerful GUI agents to determine the correct temporal order of two screenshots sampled from the same task trajectory.

\begin{figure}[t]
    \centering
    \includegraphics[width=0.8\columnwidth]{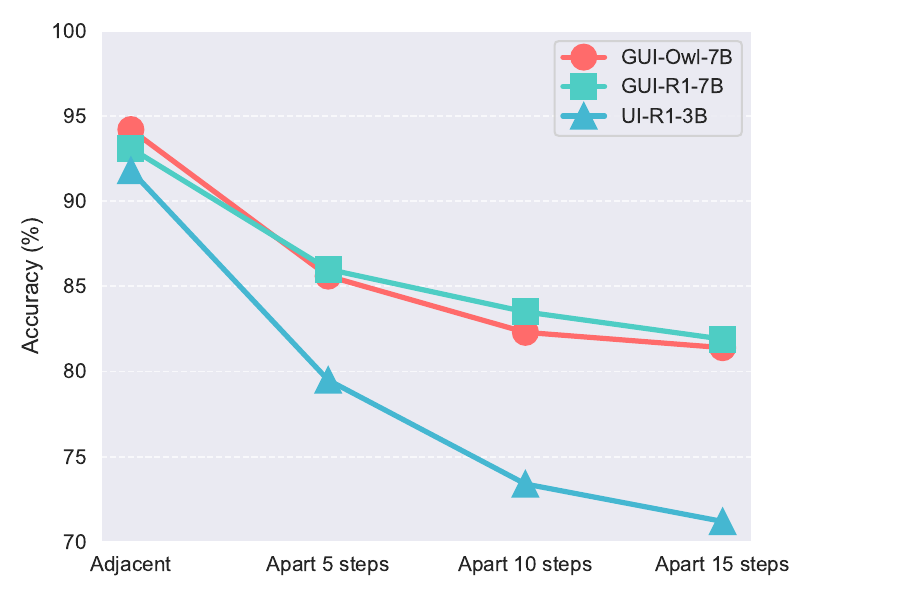} 
    \caption{Temporal Judgement Accuracy. While accuracy is high for adjacent steps, it drops dramatically as the step interval increases. This result empirically demonstrates that screenshots fail to represent task state sufficiently, and we need a mechanism to record progress for long-horizon tasks.}
    \label{fig:temporal_reasoning}
\end{figure}

Results in \Cref{fig:temporal_reasoning}, reveal a clear dichotomy. The model's accuracy is high for adjacent steps, but drops dramatically as the step interval between the screenshots increases. This failure occurs for two primary reasons: First, a long-horizon task often involves repeating screenshots, such as the Home screen or going back to an interface, leading to confusion about its true progress. Second, tasks frequently navigate into Out-of-Distribution (OOD) interfaces \cite{wu2025gem}, such as deep, specialized menus that the model has not encountered during training. When faced with an unfamiliar OOD screen, the model has no prior knowledge or context to determine its state in the overall task.

This failure highlights the critical need for a mechanism to explicitly maintain high-level semantic state. To address both this critical state awareness bottleneck and the challenge of responsibility coupling outlined in our introduction, we then introduce our CES framework and the staged execution-feedback RL algorithm in the subsequent section.

\begin{figure*}[t]
    \centering
    \includegraphics[width=\textwidth]{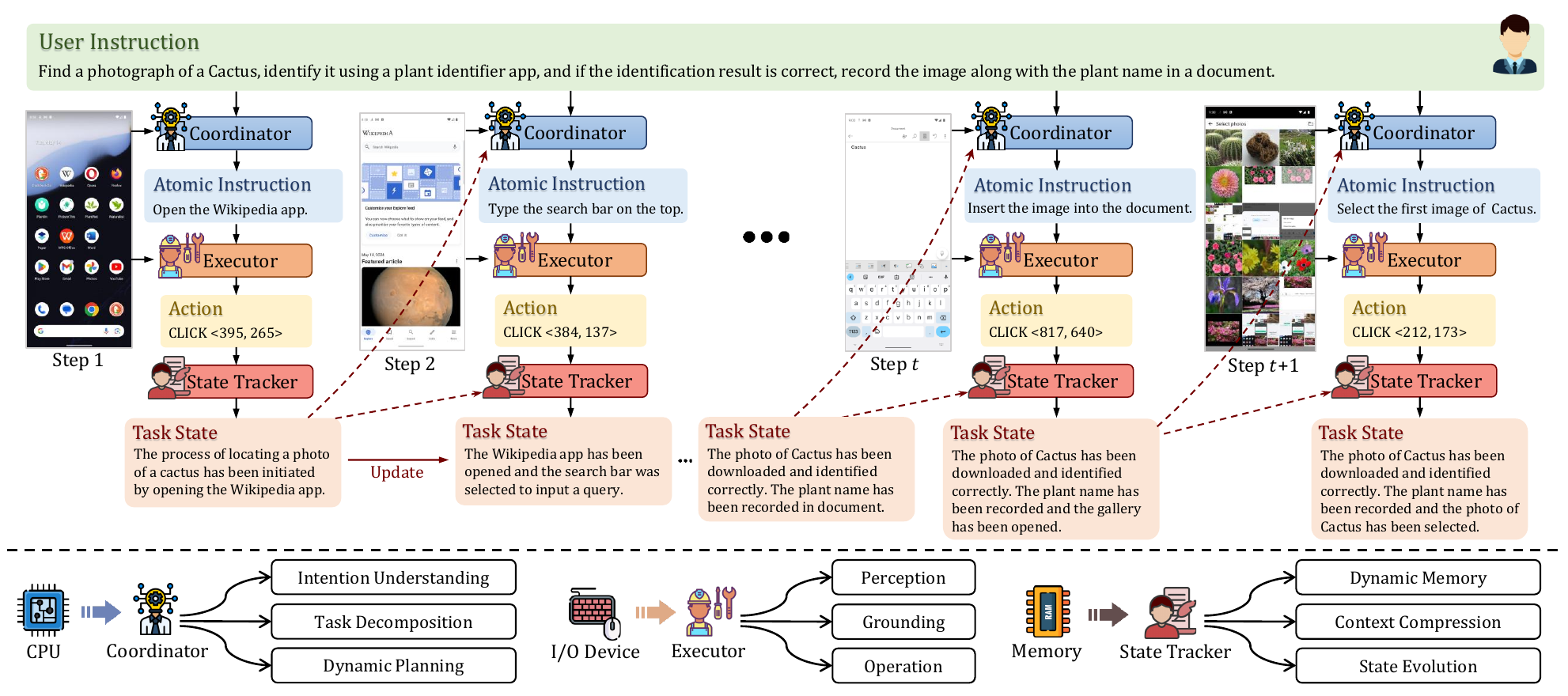} 
    \caption{The CES multi-agent loop framework. CES executes complex long-horizon tasks through the collaboration of three specialized agents. The Coordinator, as the task scheduling and decision-making core, combines the user's high-level instruction and the current task state (provided by the State Tracker) to decompose the task into a clear atomic instruction. The Executor, acting as the tool, precisely executes this atomic instruction and interacts with the GUI environment. Finally, the State Tracker, as the memory, observes the Executor's output and updates it into a high-semantic task state summary, which is then fed back to the Coordinator for the next step of decision-making.}
    \label{fig:ces_framework}
\end{figure*}

\section{Methodology}
\subsection{Task Setup}
We define the GUI task as a Markov Decision Process (MDP), characterized by the tuple $\langle \mathcal{S}, \mathcal{A}, \mathcal{F}, R \rangle$, where $\mathcal{S}$ is the state space containing all possible GUI environment states; $\mathcal{A}$ is the action space representing all executable atomic actions; $\mathcal{F}: \mathcal{S} \times \mathcal{A} \rightarrow \mathcal{S}$ is the state transition function; and $R: \mathcal{S} \times \mathcal{A} \rightarrow [0, 1]$ is the reward function. At each timestep $t$, the agent's policy $\pi$ generates a chain of thought $th^t$ and an action $a^t \in \mathcal{A}$ based on the user-specified natural language instruction $q$, the current screen state $s^t \in \mathcal{S}$, and the historical action information $H^t$, that is $(th^t, a^t) \sim \pi(\cdot | q, s^t, H^t)$. Our goal is to learn an optimal policy $\pi^*$ that maximizes the cumulative reward. However, directly learning an end-to-end policy $\pi$ faces significant challenges. Therefore, we decompose this decision-making process into a structured multi-agent collaborative loop, transforming the original MDP problem into a series of more manageable sub-problems.

\subsection{CES Collaborative Loop}
As shown in \Cref{fig:ces_framework}, our CES framework operates as a loop information flow: the Coordinator plans based on the state from the State Tracker, the Executor executes the instruction, the State Tracker updates the state based on the execution result, and the Coordinator receives the new state for the next planning step. This division of labor is inspired by a modern operating system, where the Coordinator acts as the ``CPU'' (planning), the Executor as the ``I/O Device'' (action), and the State Tracker as the ``Dynamic Memory'' (state management). Through this OS-inspired design, we completely separate high-level strategic planning from low-level precise execution, allowing each agent to focus on its core capabilities and jointly accomplish complex long-horizon tasks through efficient collaboration.

\subsubsection{Coordinator}
The Coordinator is the strategic decision-making core of the task. Its primary responsibility is to translate the user's high-level, often ambiguous, natural language instructions into a series of clear, executable steps. To make decisions that align with the long-term strategy yet adapt to the current situation, the Coordinator fuses three types of information at each timestep $t$: the user's high-level instruction $q$, which serves as the global macro-objective, the compressed high-semantic state summary $m^{t-1}$ provided by the State Tracker, and the current screenshot $s^t$. Formally,
\begin{equation}
    l^t = \pi_c(q, m^{t-1}, s^t),
\end{equation}
where $\pi_c$ denotes the Coordinator agent's policy, and $l^t$ is the generated atomic instruction, which focuses on guiding the Executor on the specific action to perform in the current environment. Furthermore, when exceptions or errors occur, the Coordinator can reflect and re-plan the task, guiding the Executor to take corrective actions and return to the correct path.

\subsubsection{Executor}
The Executor is the action endpoint in the CES framework. Its role is strictly limited to translating the atomic instruction $l^t$ from the Coordinator into a physical operation on the interface. The core advantage of this design is the complete separation of cognitive load: the Executor does not need to understand the user's long-term intent or maintain complex task context. Its sole task is to find the target element described by $l^t$ on the current screen $s^t$ and perform the corresponding action. Formally,
\begin{equation}
    u^t = (th^t, a^t) = \pi_e(l^t, s^t),
\end{equation}
where $u^t$ is the Executor's output, including the chain of thought $th^t$ and the standardized action $a^t$, and $\pi_e$ is the Executor's policy.

\subsubsection{State Tracker}
The State Tracker is key to solving the long-horizon challenges for GUI agents. Its core function is dynamic context compression and state updating. It acts as the framework's dynamic memory unit, shifting the agent's state understanding task from processing high-dimensional, redundant historical screenshots to a low-dimensional, high-semantic natural language space. It is a language model that does not directly perceive the GUI environment but rather infers and generates the new state summary $m^t$ by understanding the Executor's output $u^t$, combined with the user intent $q$ and the previous task state $m^{t-1}$:
\begin{equation}
    m^t = \pi_s(q, m^{t-1}, u^t),
\end{equation}
where $\pi_s$ is the State Tracker's policy. This natural language-based state evolution not only compresses information and filters visual noise but also provides an exceptionally clear and coherent basis for the Coordinator's decisions, effectively solving context loss in long-horizon tasks.

\subsection{Staged Execution-Feedback Reinforcement Learning}
We post-train the Coordinator on vision-language model and post-train the State Tracker on language model. Notably, in our framework, the Executor is designed as a frozen, swappable component. We do not train it; instead, we directly use a fixed, powerful pretrained GUI model. This design allows us to focus our research on optimizing the Coordinator's planning and the State Tracker's state understanding, avoiding interference from low-level execution details. This also demonstrates the generality and plug-and-play nature of the CES framework, which can be combined with any powerful execution model as its brain to enhance long-horizon task capabilities.

\begin{figure}[t!]
    \centering
    \includegraphics[width=0.7\columnwidth]{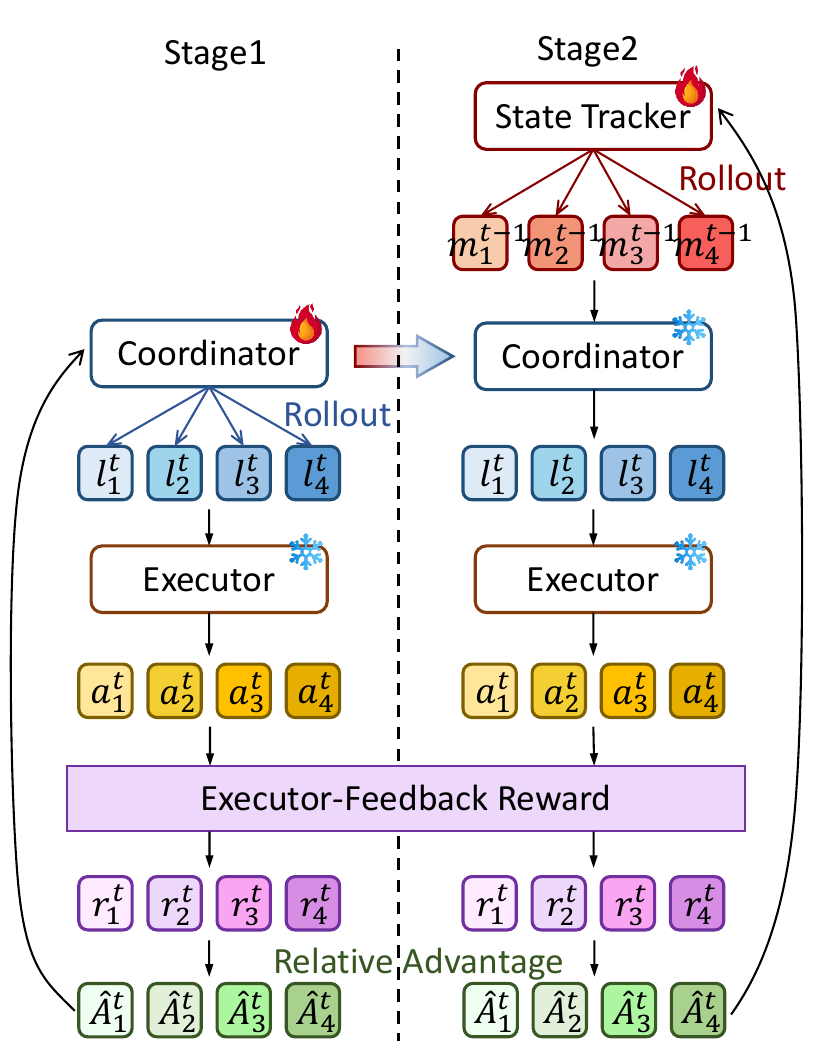} 
    \caption{Our proposed staged execution-feedback RL strategy. This strategy utilizes the Execution-Feedback Reward from a fixed Executor to sequentially optimize the Coordinator (Stage 1) and State Tracker (Stage 2) in two independent training phases.}
    \label{fig:staged_rl}
\end{figure}

\subsubsection{Warm-up SFT}
Firstly, we perform warm-up SFT for the Coordinator and State Tracker. This aims to let each agent learn its basic role, responsibilities, and strict output format. We use trajectories from existing datasets and, through automated scripts, we construct high-quality (input, output) data pairs for the preliminary fine-tuning of both agents.

\subsubsection{Rule-based RL}
\paragraph{GRPO.} To further enhance the model's generalization, we use RL to optimize the models, learning a planning policy that maximizes long-term task success. We use the GRPO algorithm \cite{shao2024deepseekmath}: 
\begin{equation}
\begin{split}
\mathcal{J}(\theta) = & \mathbb{E}\left[\frac{1}{N}\sum_{i=1}^{N}\min(\rho_i(\theta)\hat{A}_i, \text{clip}(\rho_i(\theta), 1\pm\epsilon)\hat{A}_i)\right] \\
& - \beta D_{KL}(\pi_\theta || \pi_{ref}),
\end{split}
\end{equation}
where $\rho_i(\theta) = \frac{\pi_\theta(o_i)}{\pi_{\theta_{old}}(o_i)}$ is the importance sampling ratio, $\pi_\theta$ and $\pi_{\theta_{old}}$ denote the current and previous policies, and $o_i$ is the candidate output. $\epsilon$ is the clipping hyperparameter, and $\beta$ controls the KL penalty against a reference policy $\pi_{ref}$.

For each input, the model generates $N$ candidate outputs $\mathcal{O}=\{o_1, o_2, \dots, o_N\}$, and each output is scored by a rule-based reward function to get $\mathcal{R}=\{r_1, r_2, \dots, r_N\}$. The estimated relative advantage $\hat{A}_i$ is then calculated as:
\begin{equation}
\hat{A}_i = \frac{r_i - \text{mean}(\{r_1, r_2, \dots, r_N\})}{\text{std}(\{r_1, r_2, \dots, r_N\})}.
\end{equation}

\paragraph{Execution-Feedback Reward.} Assigning rewards for agents responsible for abstract tasks (like planning) is difficult. Directly evaluating the quality of the Coordinator's atomic instructions or the State Tracker's state summaries lacks an objective standard and cannot guarantee their contribution to the final task success. To address this challenge, we do not directly evaluate the intermediate outputs. Instead, these outputs are passed through the CES loop to the Executor. The action output by the Executor is then objectively scored using a verifiable, rule-based reward function. This reward signal, originating from the final execution, is back-propagated to optimize the policies of the Coordinator and State Tracker.

\begin{table*}[t]
    \caption{Long-horizon task performance comparison on three benchmarks. \textbf{Bold} indicates the best, \underline{underline} indicates the second best. * indicates that the data of the benchmark is from GUI-R1's original paper.}
    \label{tab:main_results}
    \centering
    \resizebox{0.85\textwidth}{!}{%
    \begin{tabular}{@{}l|c|ccc|ccc|ccc@{}}
    \toprule
    \multirow{2}{*}{\textbf{Model}} & \multirow{2}{*}{\textbf{Method}} & \multicolumn{3}{c|}{\textbf{AITZ}} & \multicolumn{3}{c|}{\textbf{AMEX}} & \multicolumn{3}{c}{\textbf{GUI-Odyssey}} \\
     & & Type & GR & SR & Type & GR & SR & Type & GR & SR \\
    \midrule
    GPT-4o & Zero Shot & -- & -- & -- & -- & -- & -- & 37.50 & 14.17 & 5.36\textsuperscript{*}\\
    Qwen2.5-VL-7B & Zero Shot & 34.23 & 55.27 & 18.11 & 59.52 & 48.24 & 35.10 & 55.60 & 37.78 & 34.37\textsuperscript{*}\\
    OS-Atlas-7B & SFT & 38.52 & 44.14 & 25.97 & 55.11 & 40.30 & 33.89 & 60.42 & 39.74 & 26.96\textsuperscript{*}\\
    UI-R1-3B & RL & 41.63 & 49.27 & 24.55 & 60.23 & 41.78 & 35.81 & 52.16 &  34.46 & 32.49\textsuperscript{*}\\
    GUI-Owl-7B & RL & 53.86 & 52.08 & 32.70& 61.56 & 48.38 & 40.48 & 60.60 & 45.96 & 35.82\\
     
    SWIRL & Multi-Agent & -- & -- & -- & -- & -- & -- & \underline{74.87} & \textbf{66.39} & \underline{51.65} \\
    \midrule
     
    GUI-R1-7B & RL & 52.73 & 54.92 & 30.59 & 67.26 & \underline{57.12} & \underline{43.69} & 65.49 & 43.64 & 38.79\textsuperscript{*}\\
     
     \rowcolor{lightgray}+ GPT-5 & Multi-Agent & \underline{62.50} & \underline{59.10} & \underline{40.55} & \underline{72.80} & 52.15 & 35.80 & 70.37 & 49.92 & 42.47 \\
     
    \rowcolor{midgray} \textbf{+ CES (Ours)} & Multi-Agent & \textbf{64.44} & \textbf{64.58} & \textbf{43.05} & \textbf{77.57} & \textbf{61.64} & \textbf{48.48} & \textbf{79.24} & \underline{63.82} & \textbf{53.69} \\
    \bottomrule
    \end{tabular}%
    }
\end{table*}

\begin{table*}[t]
    \caption{Main results evaluating CES framework's Effectiveness and Generalization. \textbf{Baseline}: The model is used directly. \textbf{CES-P}: The same base model acts as all three roles within the CES framework via Prompting. \textbf{CES}: Our full framework, using our specialized Coordinator and State Tracker, with the base model serving as the Executor. \textbf{Bold} indicates the best.}
    \label{tab:main_experiment_ces}
    \centering
    \resizebox{\textwidth}{!}{%
    \begin{tabular}{@{}ll|ccc|ccc@{}}
    \toprule
    \multirow{2}{*}{\textbf{Model}} & \multirow{2}{*}{\textbf{Setting}} & \multicolumn{3}{c|}{\textbf{AMEX}} & \multicolumn{3}{c}{\textbf{GUI-Odyssey}} \\
    & & Type & GR & SR & Type & GR & SR \\
    \midrule
    & Baseline & 60.23 & 41.78 & 35.81 & 52.16 & 34.46 & 32.49 \\
    \rowcolor{lightgray} \cellcolor{white} &  CES-P & 42.52 (-17.71) & 52.15 (+10.37) & 29.12 (-6.69) & 25.81 (-26.35) & 53.05 (+18.59) & 14.44 (-18.05) \\
    \rowcolor{midgray} \cellcolor{white} \multirow{-3}{*}{UI-R1-3B} & \textbf{CES} & \textbf{70.39 (+10.16)} & \textbf{66.28 (+24.50)} & \textbf{43.38 (+7.57)} & \textbf{66.37 (+14.21)} & \textbf{64.29 (+29.83)} & \textbf{38.04 (+5.55)} \\
    \midrule
     & Baseline & 61.56 & 48.38 & 40.48 & 60.60 & 45.96 & 35.82 \\
    \rowcolor{lightgray} \cellcolor{white} & CES-P & 69.18 (+7.62) & 53.05 (+4.67) & 44.91 (+4.43) & 66.03 (+5.43) & 52.15 (+6.19) & 37.53 (+1.71) \\
    \rowcolor{midgray} \cellcolor{white} \multirow{-3}{*}{GUI-Owl-7B} & \textbf{CES} & \textbf{75.72 (+14.16)} & \textbf{61.19 (+12.81)} & \textbf{47.24 (+6.76)} & \textbf{74.87 (+14.27)} & \textbf{61.39 (+15.43)} & \textbf{46.65 (+10.83)} \\
    \midrule
     & Baseline & 69.13 & 53.24 & 43.16 & 67.15 & 47.33 & 39.60 \\
    \rowcolor{lightgray} \cellcolor{white} & CES-P & \textbf{81.02 (+11.89)} & \textbf{63.80 (+10.56)} & \textbf{56.34 (+13.18)} & 76.90 (+9.75) & 65.10 (+17.77) & 53.88 (+14.28) \\
    \rowcolor{midgray} \cellcolor{white} \multirow{-3}{*}{GUI-Owl-32B}& \textbf{CES} & 78.55 (+9.42) & 63.11 (+9.87) & 52.05 (+8.89) & \textbf{79.58 (+12.43)} & \textbf{65.42 (+18.09)} & \textbf{56.75 (+17.15)} \\
    \bottomrule
    \end{tabular}%
}
\end{table*}

Specifically, we design a rule-based reward function, termed the \textbf{Execution-Feedback Reward}:
\begin{equation}
R = \alpha_1 R_{format} + \alpha_2 R_{executor},
\end{equation}
where $R_{format}$ is the format reward, rewarding the model for outputting in the $<$think$>$ and $<$answer$>$ tags, $R_{executor}$ is the executor reward, and $\alpha_1, \alpha_2$ are coefficients. By passing the Coordinator's output and the current screenshot $s^t$ to the executor, we derive the executor reward from its action:
\begin{equation}
R_{executor} =  \gamma_1 R_{type} + \gamma_2 R_{param},
\end{equation}
where $\gamma_1, \gamma_2$ are coefficients, $R_{type}$ rewards the correct action type, and $R_{param}$ rewards the correct action parameters. More details can be found in Appendix A.

\paragraph{Staged Optimization.} Based on the GRPO algorithm and the Execution-Feedback Reward, we optimize the Coordinator and State Tracker in two distinct stages, as shown in \Cref{fig:staged_rl}. This strategy ensures that high-level planning and state tracking are always optimized towards goals that are verifiably effective and easily understood by the Executor.

Stage 1: Optimizing the Coordinator's Planning Capability. In this stage, we use a frozen Executor model and calculate the reward based on its execution results to update the Coordinator's policy network. Since we do not yet have a trained State Tracker, the required task state $m^{t-1}$ is sourced directly from the ground-truth annotated states in our preprocessed dataset. This ensures the Coordinator focuses on learning the mapping from a ``perfect" state to the optimal atomic instruction.

Stage 2: Optimizing the State Tracker's State Evolution Capability. After the Coordinator is trained, we freeze its parameters. In this stage, the State Tracker's capability is the focus of optimization. The task state $m^{t-1}$ it generates is passed through the fixed Coordinator and Executor, influencing the entire decision chain. The final Execution-Feedback Reward from the Executor is used exclusively to optimize the State Tracker, teaching it to generate the most valuable state information that helps the Coordinator make optimal decisions. In this way, the State Tracker is explicitly trained to produce state summaries that are maximally useful to the fixed Coordinator's policy, effectively learning to generate what the Coordinator understands best.

\section{Experiments}
\subsection{Settings}
\paragraph{Training Details.} We choose Qwen2.5-VL-7B \cite{bai2025qwen25vl} as the base model for the Coordinator and Qwen3-4B \cite{yang2025qwen3} for the State Tracker. For the warm-up SFT, we use the LLaMA Factory framework and trained for 1 epoch with a learning rate of 5e-5 to prevent overfitting. For RL, we use the Verl framework to train the Coordinator for 10 epochs with a learning rate of 1e-6 and State Tracker for 5 epochs. We use GUI-R1-7B \cite{luo2025guir1} as the Executor to calculate the reward function when training. For the reward coefficients, we set $\alpha_1, \alpha_2$ to 0.1 and 0.9, respectively, and $\gamma_1, \gamma_2$ to 0.2, and 0.8, respectively. All experiments are conducted on 8 $\times$ 80G GPUs.

\paragraph{Benchmarks.} To thoroughly evaluate our framework's ability to solve long-horizon tasks, we leverage three benchmarks for long-horizon, complex tasks, including AITZ \cite{zhang2024android}, AMEX \cite{chai2025amex} and GUI-Odyssey \cite{lu2024guiodyssey}, with an average steps of 7.5, 12.8 and 15.3 respectively. We only use the high-level instructions provided by the datasets, not the low-level ones.

\paragraph{Evaluation Metrics.} Following previous work \cite{lu2025swirl, luo2025guir1, wu2024osatlas}, we use three common metrics for GUI agent evaluation: (i) \textbf{Type}, the prediction accuracy of the action type. (ii) \textbf{GR} (Grounding), the click point prediction accuracy, where a prediction is considered correct if the point falls within the ground-truth bounding box. (iii) \textbf{SR} (Success Rate), correct only if both action type and parameters are correct. Parameter correctness includes: the point being in the bounding box, correct scroll direction, or an F1 similarity $> 0.5$ for input text.

\subsection{Main Results}

\subsubsection{Long-Horizon Task Performance} As shown in \Cref{tab:main_results}, CES achieves compelling performance on complex GUI tasks by implementing atomic instruction and progress state management. On top of the GUI-R1-7B executor baseline, our method improves the Type accuracy by an average of 10.38\% across all benchmarks. This significant gain strongly validates the effectiveness of our decoupled design: by having the Coordinator bear the cognitive load of high-level planning and generate clear atomic instructions, the Executor no longer needs to reason about complex task context and can focus on simple perception and localization, thereby greatly enhancing its execution accuracy and stability.

Furthermore, when we use GPT-5 via prompting as the Coordinator and State Tracker, the performance only shows a minor improvement, exemplified by an average increase of only 4\% in Type accuracy, while some metrics even degrade. This indicates that while powerful general-purpose models can offer some improvement through prompt engineering, the effect is unstable. In contrast, our CES framework, through targeted execution-feedback reinforcement learning, enables the Coordinator and State Tracker to learn planning and state understanding strategies that are much better aligned with GUI tasks, thus achieving stable and significant performance superiority.

\subsubsection{Effectiveness and Generalization}

To validate the efficiency and generality of our CES framework, we conducted a further experiment using three models of varying sizes. For each model, we tested three distinct configurations: 
(i) \textbf{Baseline}: The model is used directly.
(ii) \textbf{CES-P}: The same base model acts as all three roles within the CES framework via prompting. 
(iii) \textbf{CES}: Our full framework, using our specialized Coordinator and State Tracker, with the base model serving as the Executor. 
We conducted extensive evaluations on two AMEX and GUI-Odyssey with long average task steps, with the results shown in Table \ref{tab:main_experiment_ces}.

From these results, we derive several key insights:

(i) \textbf{Architectural Superiority: Solving State Awareness and Cognitive Load.}
A consistent finding across all model sizes is that the CES-P setup significantly outperforms the baseline. The baseline model fails in long-horizon tasks because it lacks a mechanism to manage task state, forcing it to infer progress from ambiguous screenshots. The CES framework solves this by introducing a State Tracker, which provides explicit, high-semantic memory, and a Coordinator, which decouples the cognitive load of planning from the immediate burden of execution.

(ii) \textbf{Validating the Capability Conflict Hypothesis.}
While CES-P helps all models, the magnitude of the improvement reveals evidence of capability conflict while training. For the small UI-R1-3B, the CES-P setup actually resulted in a performance degradation, with a significant drop of 18.05\% in SR on the GUI-Odyssey. In contrast, the large GUI-Owl-32B sees a massive improvement of 13.18\% on AMEX. 
This suggests that the 3B model, due to its limited parameter capacity, struggled to learn these distinct planning and execution capabilities. The training-time conflict was too severe for it to be overcome. Conversely, the 32B model learned planning, execution, and state understanding abilities during training, even though these abilities remained coupled. The CES-P setup reveals the latent high-level skills that the 32B model had already acquired tentatively. This validates our hypothesis that it is extremely difficult to train these conflicting capabilities within a single policy network, and this challenge is magnified in models with smaller parameter counts.

(iii) \textbf{The Efficacy of Specialized, Trained Components.}
Comparing the CES-P to CES highlights the significant value of our training strategy. For instance, on GUI-Odyssey, our specialized training boosts the GUI-Owl-32B's SR from 39.60\% to 56.75\%, and the GUI-Owl-7B's SR from 37.53\% to 46.65\%. This demonstrates that our Coordinator and State Tracker via staged execution-feedback RL are highly effective and vastly superior to simply prompting a non-specialized model for these roles. Even, we can get the comparable effect as the Gui-Owl-32B model just by using 7B Coordinator and 4B State Tracker. This also proves the effectiveness of decoupling training.

(iv) \textbf{Generality of the CES Framework.}
Finally, our full framework, CES, provides a substantial and consistent performance improvement over the baseline for every executor model tested. This confirms that CES is a robust and generalizable plug-and-play solution that can effectively enhance any underlying executor, dramatically improving its long-horizon task automation capabilities.

\subsection{Analysis}
\paragraph{Component and Training Ablation.} We removed the Coordinator, State Tracker, and the RL stage from our framework to observe the effects. As shown in \Cref{tab:ablation_components}, when we remove the Coordinator and feed the user's high-level instruction directly to the Executor, performance drops significantly, such as SR drops by 12.77\% on GUI-Odyssey. This indicates that the Executor still struggles to understand high-level instructions and make current decisions, proving the necessity of decoupling planning from execution. When the State Tracker is removed, where we instead record the last four action histories as most methods and input them to the Coordinator, performance also drops significantly. This shows the critical role of information compression and state management in long-horizon tasks; without the concise, high-semantic summary from the State Tracker, the Coordinator cannot make correct plans. Finally, when we test the model using only SFT warm-up, its performance is far below the final RL-optimized model. This demonstrates that while SFT allows agents to learn basic roles and output formats, execution-feedback RL is an indispensable step to acquire a generalizable, optimal planning policy.

\begin{table}[t!]
    \caption{Ablation study on components and training stages.}
    \label{tab:ablation_components}
    \centering
    \resizebox{\columnwidth}{!}{%
    \begin{tabular}{@{}l|ccc|ccc@{}}
    \toprule
    \multirow{2}{*}{\textbf{Model}} & \multicolumn{3}{c|}{\textbf{AMEX}} & \multicolumn{3}{c}{\textbf{GUI-Odyssey}} \\
 & Type      & GR      & SR      & Type       & GR       & SR       \\ \midrule
    CES                    & \textbf{77.57}     & \textbf{61.64}   & \textbf{48.48}   & \textbf{79.24}      & \textbf{63.82}    & \textbf{53.69}    \\
    w/o Coordinator                & 62.63     & 50.57   & 33.27   & 70.11      & 48.03    & 39.15    \\
    w/o State Tracker              & 70.62     & 55.70   & 42.08   & 73.34      & 45.10    & 42.52    \\
    w/o RL (SFT only)              & 72.37     & 53.47   & 36.54   & 72.18      & 51.33    & 42.89    \\ \bottomrule
    \end{tabular}%
    }
\end{table}

\begin{figure}[t]
    \centering
    \includegraphics[width=\columnwidth]{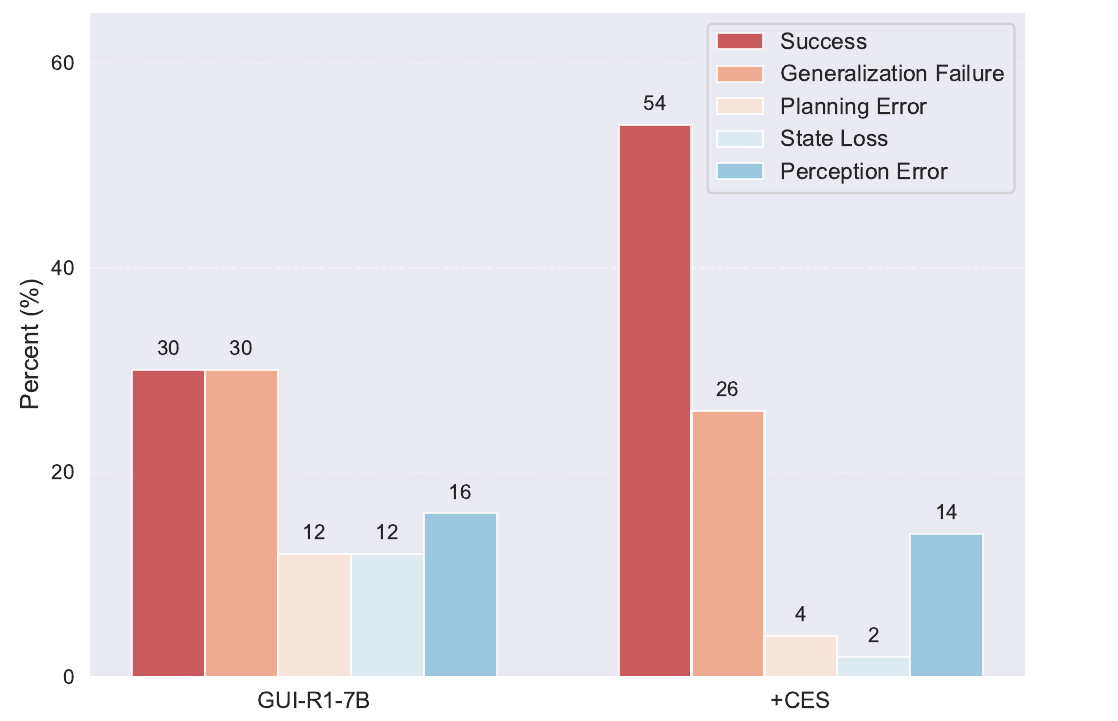}
    \caption{Failure Case Analysis. Compared to the baseline, our CES framework almost completely eliminates cognitive errors like State Loss and Planning Error.}
    \label{fig:failure_analysis}
\end{figure}

\paragraph{Failure Case Analysis.} To understand the source of our CES framework's advantages, we conducted a detailed failure case analysis, with results shown in \Cref{fig:failure_analysis}. Specifically, the CES framework almost completely eliminated State Loss errors, reducing the count from 14\% to 2\% and significantly reducing Planning Error from 12\% to 4\%. This result demonstrates that our framework's performance gain stems precisely from our specially trained Coordinator and State Tracker, which successfully solve the core challenges of planning and state management. In contrast, the errors attributable to the frozen Executor, namely Perception Error and Generalization Failure, remained largely unchanged. The performance bottleneck has now effectively shifted to the inherent perceptual limitations of  Executor itself. More case studies can be found in Appendix D.

\section{Conclusion}
In this paper, we propose a staged execution-feedback RL strategy to address the core challenges of long-horizon GUI automation. Our algorithm confronts the capability overload and lack of task progress awareness by shifting the training strategy. Instead of training a unified policy, our method thoroughly decouples high-level strategic planning from low-level precise execution. It leverages verifiable results to efficiently optimize high-level agents responsible for scheduling.
Critically, our algorithm trains a dedicated State Tracker that resolves the agent's difficulties in long-horizon tasks via dynamic context compression and high-semantic state summarization. This algorithmic approach is instantiated in our CES framework.
Extensive experiments on challenging long-horizon GUI benchmarks demonstrate that our trained high-level modules are a generalizable, plug-and-play solution that significantly enhances the long-horizon planning and state management capabilities of various Executor models. 
In the future, synergetic evolution and joint training of multi-agent system for GUI tasks may be a promising direction.

\section*{Acknowledgment}
This work was supported by the National Key R\&D Program of China (No.2024YFC3306500), National Natural Science Foundation of China (62406188), and Shanghai Municipal Science and Technology Major Project (2025SHZDZX025G08).

{
    \small
    \bibliographystyle{ieeenat_fullname}
    \bibliography{ref}
}

\newpage
\appendix
\onecolumn
\section{Details of Staged Execution-Feedback Reinforcement Learning}
\label{app:A}

\subsection{Staged Optimization}
The core idea of our algorithm is to completely decouple the training of the Coordinator, denoted as $\pi_c$, and the State Tracker, denoted as $\pi_s$. This is achieved through two independent optimization stages, each with a specifically defined objective function. We use parameters $\theta_c$ and $\theta_s$ to denote the trainable parameters of the Coordinator and State Tracker, respectively.

\subsubsection*{Stage 1: Optimizing the Coordinator}
The Coordinator's optimization objective, $\mathcal{J}(\theta_c)$, is to maximize the following expectation:

\begin{equation}
\mathcal{J}(\theta_c) = \mathbb{E}_{\substack{(q, s^t, m_{gt}^{t-1}, a_{gt}^t) \sim \mathcal{D} \\ \mathcal{L} \sim \pi_{\theta_{c,old}}}} 
\left[ \frac{1}{N}\sum_{i=1}^{N}\min(\rho_{i}^{t,(c)}(\theta_c)\hat{A}_{i}^{t,(c)}, \text{clip}(\rho_{i}^{t,(c)}(\theta_c), 1\pm\epsilon)\hat{A}_{i}^{t,(c)}) \right] - \beta D_{KL}(\pi_{\theta_c} || \pi_{c, ref}).
\end{equation}

We firstly sample $N$ candidate atomic instructions $\mathcal{L} = \{l_1^t, \dots, l_N^t\}$ using the old Coordinator policy $\pi_{\theta_{c,old}}$:
\begin{equation}
    l_i^t \sim \pi_{\theta_{c,old}}(\cdot | q, m_{gt}^{t-1}, s^t),
\end{equation}    
where $q, m_{gt}^{t-1}, s^t$ are sampled from the dataset $\mathcal{D}$. For each candidate instruction $l_i^t$, obtain the executed action $a_i^t$ via the frozen Executor $\pi_e$:
\begin{equation}
    a_i^t = \pi_e(l_i^t, s^t).
\end{equation}
The $a_i^t$ are used to calculate reward as:
\begin{equation}
    r_i^{t,(c)} = \alpha_1 R_{format}(l_i^t) + \alpha_2 R_{executor}(a_i^t, a_{gt}^t).
\end{equation}
Based on the rewards of the $N$ candidates $\{r_1^{(c)}, \dots, r_N^{(c)}\}$, calculate the relative advantage:
\begin{equation}
    \hat{A}_{i}^{t,(c)} = \frac{r_{i}^{t,(c)} - mean(\{r_1^{t,(c)}, r_2^{t,(c)},...,r_N^{t,(c)}\})}
    {std(\{r_1^{t,(c)}, r_2^{t,(c)},...,r_N^{t,(c)}\})}.
\end{equation}

At the same time, we calculate the probability ratio between the new Coordinator policy $\pi_{\theta_c}$ and the old policy $\pi_{\theta_{c,old}}$:
\begin{equation}
    \rho_i^{t,(c)}(\theta_c) = \frac{\pi_{\theta_c}(l_i^t | q, m_{gt}^{t-1}, s^t)}{\pi_{\theta_{c,old}}(l_i^t | q, m_{gt}^{t-1}, s^t)}.
\end{equation}
After calculating the above items, we can calculate the gradient of $\mathcal{J}(\theta_c)$ and update the strategy $\theta_c$.

\subsubsection*{Stage 2: Optimizing the State Tracker}

With $\pi_{c}$ and $\pi_{e}$ frozen, train the State Tracker $\pi_{s}$ to generate the  state $m^{t}$ that guides the fixed $\pi_{c}$ to make an optimal decision .

The State Tracker's optimization objective, $\mathcal{J}(\theta_{s})$, is to maximize the following expectation :
\begin{equation}
\mathcal{J}(\theta_{s}) = \mathbb{E}_{\substack{(q,s^{t},m_{gt}^{t-1},u_{gt}^{t-1},a_{gt}^{t})\sim\mathcal{D} \\ \mathcal{M} \sim \pi_{\theta_{s,old}}}} \left[ \frac{1}{N}\sum_{i=1}^{N}\min(\rho_{i}^{t, (s)}(\theta_s)\hat{A}_{i}^{t, (s)}, \text{clip}(\rho_{i}^{t, (s)}(\theta_s), 1\pm\epsilon)\hat{A}_{i}^{t, (s)}) \right] - \beta D_{KL}(\pi_{\theta_s} || \pi_{s, ref}).
\end{equation}

We firstly sample $N$ candidate state summaries $\mathcal{M}=\{m_{1}^{t},...,m_{N}^{t}\}$ using the old State Tracker policy $\pi_{\theta_{s,old}}$:
\begin{equation}
m_{i}^{t} \sim \pi_{\theta_{s,otd}}(\cdot | q, m_{gt}^{t-1}, u_{gt}^{t-1})
\end{equation}
This begins a chained evaluation process. For each candidate state $m_{i}^{t}$, we first obtain the instruction $l_{i}^{t}$ via the \textbf{frozen} Coordinator $\pi_{c}$ :
\begin{equation}
l_{i}^{t} = \pi_{c}(q, m_{i}^{t}, s^{t}).
\end{equation}
The instruction $l_{i}^{t}$ is passed to the frozen Executor $\pi_{e}$ to obtain the action $a_{i}^{t}$ :
\begin{equation}
a_{i}^{t} = \pi_{e}(l_{i}^{t}, s^{t}).
\end{equation}
Finally, $a_{i}^{t}$ is used to calculate the reward $r_{i}^{t, (s)}$ for the state $m_{i}^{t}$, which includes a format reward for the state and the final executor reward :
\begin{equation}
r_{i}^{t,(s)} = \alpha_{1}R_{format}(m_{i}^{t}) + \alpha_{2}R_{executor}(a_{i}^{t}, a_{gt}^{t}).
\end{equation}
Based on the rewards of the N candidates $\{r_{1}^{t, (s)},...,r_{N}^{t, (s)}\}$ , we calculate the relative advantage :
\begin{equation}
\hat{A}_{i}^{t, (s)} = \frac{r_{i}^{t,(s)} - mean(\{r_1^{t,(s)}, r_2^{t,(s)},...,r_N^{t,(s)}\})}
    {std(\{r_1^{t,(s)}, r_2^{t,(s)},...,r_N^{t,(s)}\})}.
\end{equation}
At the same time, we calculate the probability ratio between the new State Tracker policy $\pi_{\theta_{s}}$ and the old policy $\pi_{\theta_{s,old}}$ :
\begin{equation}
\rho_{i}^{t, (s)}(\theta_{s}) = \frac{\pi_{\theta_{s}}(m_{i}^{t} | q, m_{gt}^{t-1}, u_{gt}^{t-1})}{\pi_{\theta_{s,old}}(m_{i}^{t} | q, m_{gt}^{t-1}, u_{gt}^{t-1})}.
\end{equation}

\subsection{Reward Function}

The total reward $R$ is defined as a weighted sum of the format reward and the executor reward:
\begin{equation}
R = \alpha_1 R_{format} + \alpha_2 R_{executor}.
\end{equation}

The $R_{format}$ component is a binary reward that encourages the model to generate outputs in the specified format.
\begin{equation}
    R_{format}(o) = \mathbb{I}(\text{CheckFormat}(o)),
\end{equation}
where $\text{CheckFormat}(\cdot)$ is a function that returns 1 if the output $o$ strictly adheres to the $\langle\text{think}\rangle$ and $\langle\text{answer}\rangle$ tag format, and 0 otherwise.

$R_{executor}$ evaluates the contribution of the high-level agent's decision to the downstream Executor's actual execution result, based on the $i$-th candidate's action $a_i^t$ and the ground-truth action $a_{gt}^t$. It is defined as a weighted sum of a type reward and a parameter reward:
\begin{equation}
    R_{executor}(a_i^t, a_{gt}^t) = \gamma_1 R_{type}(a_i^t, a_{gt}^t) + \gamma_2 R_{param}(a_i^t, a_{gt}^t),
\end{equation}
where $R_{type}$ evaluates if the Executor's predicted action type matches the ground-truth. Let $a_i^{t, type}$ be the predicted action type and $a_{gt}^{t, type}$ be the ground-truth action type:
\begin{equation}
    R_{type}(a_i^t, a_{gt}^t) = \mathbb{I}(a_i^{t, type} = a_{gt}^{t, type}) = 
    \begin{cases} 
        1 & \text{if } a_i^{t, type} = a_{gt}^{t, type} \\ 
        0 & \text{otherwise} 
    \end{cases}
    \label{eq:reward_type}
\end{equation}
where $\mathbb{I}(\cdot)$ is the indicator function.

Parameter Reward $R_{param}$ evaluates the correctness of the predicted action parameters, consistent with the Success Rate (SR) metric. Its definition depends on the predicted action type $a_i^{t, type}$:
\begin{equation}
    R_{param}(a_i^t, a_{gt}^t) = 
    \begin{cases} 
        \mathbb{I}(p_i^{t, param} \in p_{gt}^{t, bbox}) & \text{if } a_i^{t, type} \in \{\text{`click', `long\_press'}\} \\ 
        \mathbb{I}(F1(p_i^{t, param}, p_{gt}^{t, text}) > 0.5) & \text{if } a_i^{t, type} = \text{`type'} \\
        \mathbb{I}(p_i^{t, param} = p_{gt}^{t, dir}) & \text{if } a_i^{t, type} = \text{`scroll'} \\
        \mathbb{I}(a_i^{t, type} = a_{gt}^{t, type}) & \text{otherwise} \\
    \end{cases}
    \label{eq:reward_param}
\end{equation}
where $p_i^{t, param}$ is the parameters of the action, $p_{gt}^{t, bbox}$ is the ground-truth bounding box, $F1(\cdot)$ is the F1 similarity function for text. For actions without parameters (e.g., `press\_home'), the reward is based on type correctness.

\section{Experiment Details}
\subsection{Data Collection}

\paragraph{SFT}For the warm-up SFT stage, we randomly selected 1K samples from the GUI-Odyssey training set due to its rich semantic annotations. This dataset includes: \texttt{description}: A description of the current screen; \texttt{intention}: The intent of the action to be taken, similar to a chain of thought; \texttt{low\_level\_instruction}: The specific low-level instruction for the step; \texttt{context}: A summary of actions taken before the current step.
For the Coordinator, we constructed the ground truth as: \texttt{<think>description.intention</think><answer>low\_level\_instruction</answer>}.
For the State Tracker, we used the \texttt{context} of the next step as its ground truth.

\paragraph{Staged Execution-Feedback Reinforcement Learning}
For the RL stage, rich semantic annotations are no longer required; only an execution result for the reward signal is needed. We randomly selected 3K samples from the GUI-Odyssey training set to serve as the task pool. The ground truth was the fixed action and parameters for reward calculation.

\subsection{Experiment Settings}

\paragraph{SFT} We trained the Coordinator (Qwen2.5-VL-7B) and State Tracker (Qwen3-4B) using the LLaMA-Factory framework. Both models were trained for one epoch with a learning rate of 5e-5. We used LoRA for fine-tuning, with a rank of 8 and alpha of 16.

\paragraph{Staged Execution-Feedback Reinforcement Learning}
We used a total of 8 $\times$ 80G GPUs. In Stage 1, 4 GPUs were used to deploy the fixed Executor via vLLM for reward calculation. In Stage 2, 2 GPUs were used to deploy the Coordinator and Executor respectively. For the reward coefficients, we set $\alpha_1, \alpha_2$ to 0.1 and 0.9, respectively, and $\gamma_1, \gamma_2$ to 0.2, and 0.8, respectively. Detailed hyperparameters are provided in Table \ref{tab:rl_hyperparams}.
\begin{table}[h]
\centering
\caption{Hyperparameters for Staged RL Training}
\label{tab:rl_hyperparams}
\begin{tabular}{@{}ccc@{}}
\toprule
\textbf{HyperParameter} & \textbf{Coordinator} & \textbf{State Tracker} \\ \midrule
lr & 1e-6 & 1e-6 \\
epochs & 10 & 5 \\
optimizer & AdamW & AdamW \\
train\_batch\_size & 32 & 32 \\
clip\_ratio & 0.2 & 0.2 \\
rollout\_n & 4 & 4 \\
max\_prompt\_length & 8192 & 8192 \\
max\_response\_length & 256 & 512 \\ \bottomrule
\end{tabular}
\end{table}

\subsection{Benchmarks}
AMEX is a large-scale, multi-level annotated Android GUI data set, designed to provide support for general mobile GUI control agents. It is committed to providing multi-level understanding of mobile GUI, including more than 104k high-resolution screenshots and about 3000 unique complex instructions, with an average of 12.8 steps per instruction.
AITZ (Andriod in the Zoo) is a refined data set built for the Android GUI navigation field. It is the first time to connect the perception of screen layout/ui elements with the cognition of action decision-making. It contains 2504 unique instructions and 18643 screen action pairs, covering more than 70 Android Applications.
GUI-Odyssey is a comprehensive data set that focuses on cross application GUI navigation on mobile devices. This data set aims to solve the complexity of cross application workflow. Tasks usually need to integrate multiple applications and transfer context and data between applications. It contains 8334 task tracks, with an average of 15.3 steps per track, which is the longest average step in the mobile GUI navigation dataset.

\subsection{Visualization}

\begin{figure}[h!]
    \centering
    \includegraphics[width=\columnwidth]{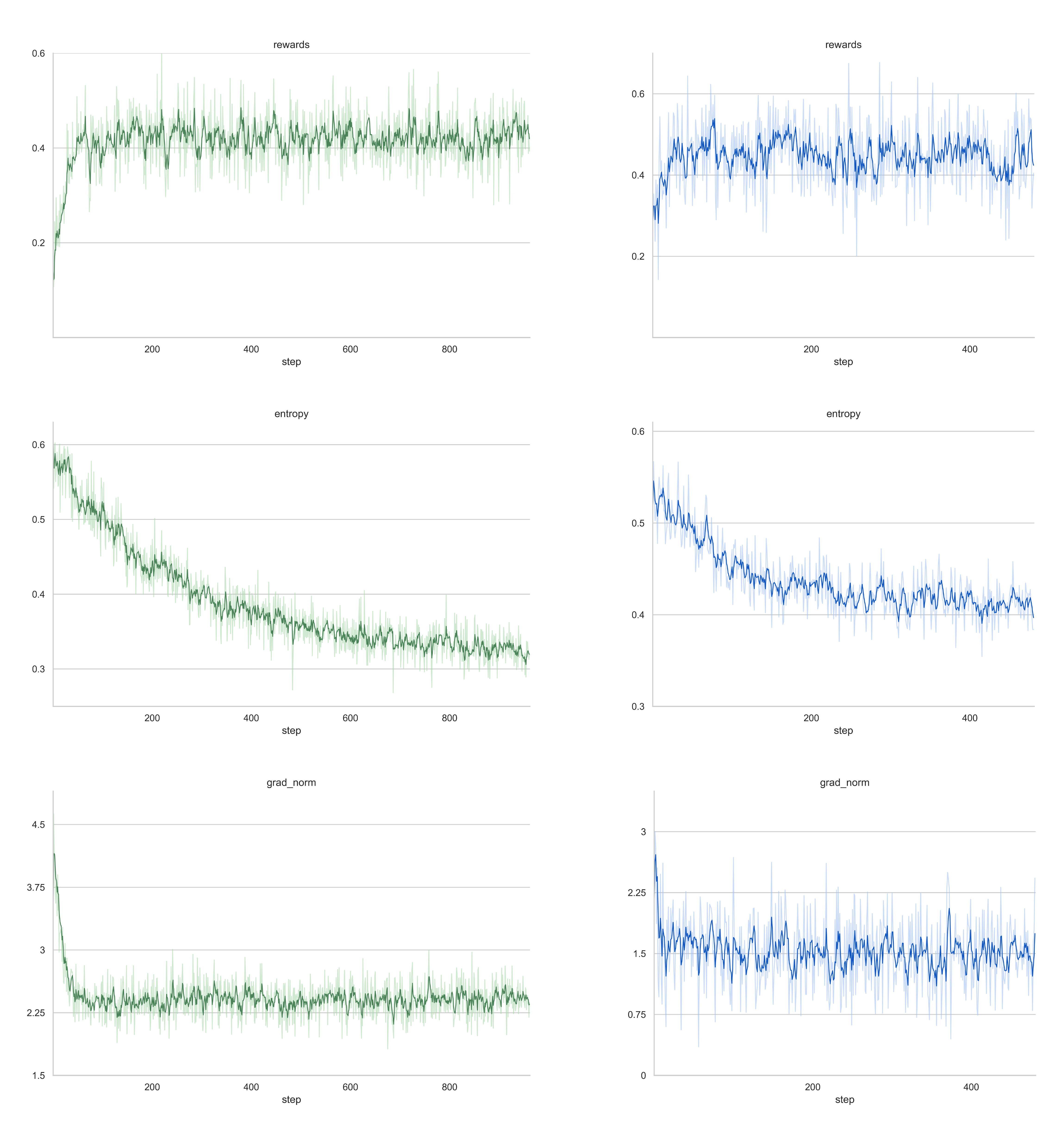}
    \caption{Coordinator (left) and State Tracker (right) training process.}
    \label{fig:vis}
\end{figure}

\Cref{fig:vis} illustrates the progression of various variables throughout the training process.

\subsection{Model Scaling Analysis}
To explore how model capacity influences the CES framework, we conducted a scaling analysis by retraining the Coordinator and State Tracker with varying parameter sizes.

\begin{figure}[h!]
    \centering
    \includegraphics[width=0.7\columnwidth]{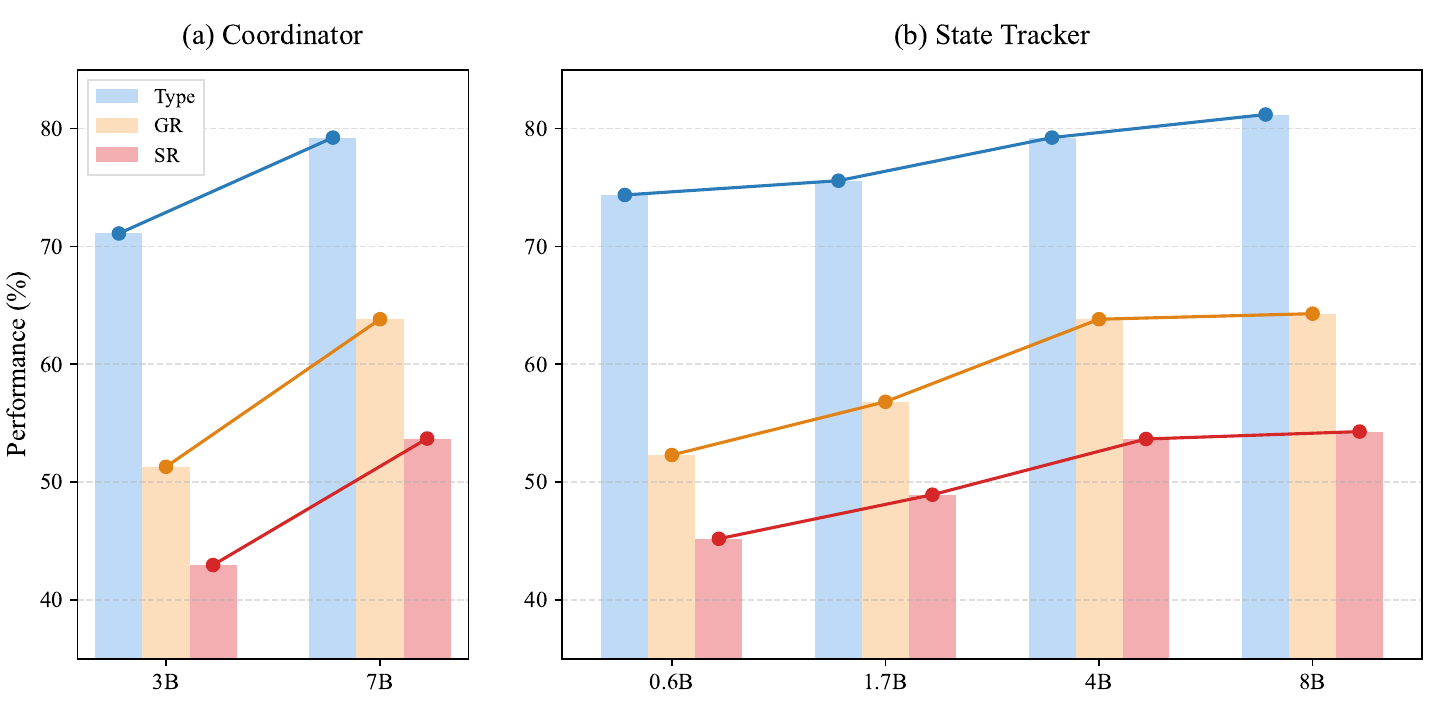}
    \caption{Model scaling analysis.}
    \label{fig:scale}
\end{figure}

For Coordinator, we investigated the trade-off between model size and planning capability by replacing the Qwen-2.5-VL-7B backbone with the compact Qwen-2.5-VL-3B. The results demonstrate a significant performance degradation across all metrics with the 3B model. This decline suggests that the smaller model lacks the capacity for complex instruction decomposition and fine-grained visual understanding required for the Coordinator role. Moreover, a weaker Coordinator creates a bottleneck that propagates to the second training stage, hindering the State Tracker's ability to learn effective state representations.

For State Tracker, we performed Qwen3 models with varying sizes (0.6B, 1.7B, 4B, and 8B) to determine the optimal configuration. As illustrated in the \Cref{fig:scale}, performance improves markedly as the model scales from 0.6B to 4B. However, increasing the size further to 8B yields only marginal gains. Therefore, we identify Qwen3-4B as the optimal choice for the State Tracker, offering a favorable balance between performance and computational efficiency.

\section{Prompt}
\label{app:c}

\begin{promptbox}{Coordinator}
You are a GUI task coordinator Agent. Your role is to actively collaborate with the Executor Agent to complete complex GUI navigation tasks. Given a high-level task description and the current state of the task, your goal is to provide a clear and precise fine-grained instruction for the Executor Agent to help accomplish the task.

Screenshot: \texttt{<image>}

High-level task: \texttt{\{high\_level\_instruction\}}

Current\_state: \texttt{\{current\_state\}}

First, think step-by-step. Put your reasoning within \texttt{<think>} tags.
After your reasoning, provide the instruction within \texttt{<answer>} tags.
\end{promptbox}

\begin{promptbox}{Executor}
You are GUI executor Agent, a reasoning GUI Agent Assistant. In this UI screenshot \texttt{<image>}, I want you to execute the command '\texttt{\{instruction\}}'.
Please provide the action to perform (enumerate from [complete, close, press home, click, press back, type, select, scroll, enter]), the point where the cursor is moved to (integer) if a click is performed, and any input text required to complete the action.

Output the thinking process in \texttt{<think> </think>} tags, and the final answer in \texttt{<answer> </answer>} tags as follows: \texttt{<think>...</think>} \texttt{<answer>}[`action': enum[complete, close, press home, click, press back, type, select, scroll, enter], `point': [x, y], `input\_text': `no input text [default]']\texttt{</answer>}
\end{promptbox}

\begin{promptbox}{State Tracker}
You are a GUI task State Tracker Agent. Your core function is dynamic context compression and state updating. You will receive the high-level user instruction, the previous task state (a summary of progress up to the last step), and the latest output of executor agent. Your task is to generate the new task state. This should be a high-semantic natural language summary that updates the previous state based on the latest action, maintaining a coherent record of the task's progress.

High-level user instruction: \texttt{\{high\_level\_instruction\}}

Latest output of executor agent: \texttt{\{executor\_output\}}

Previous Task State: \texttt{\{current\_state\}}

\end{promptbox}

These prompt are also used in +GPT-5 setting (in \Cref{tab:main_results}) and CES-P setting (in \Cref{tab:main_experiment_ces}).

\section{Case Study}
\label{app:d}

\begin{figure}[t]
    \centering
    \includegraphics[width=\columnwidth,page=3]{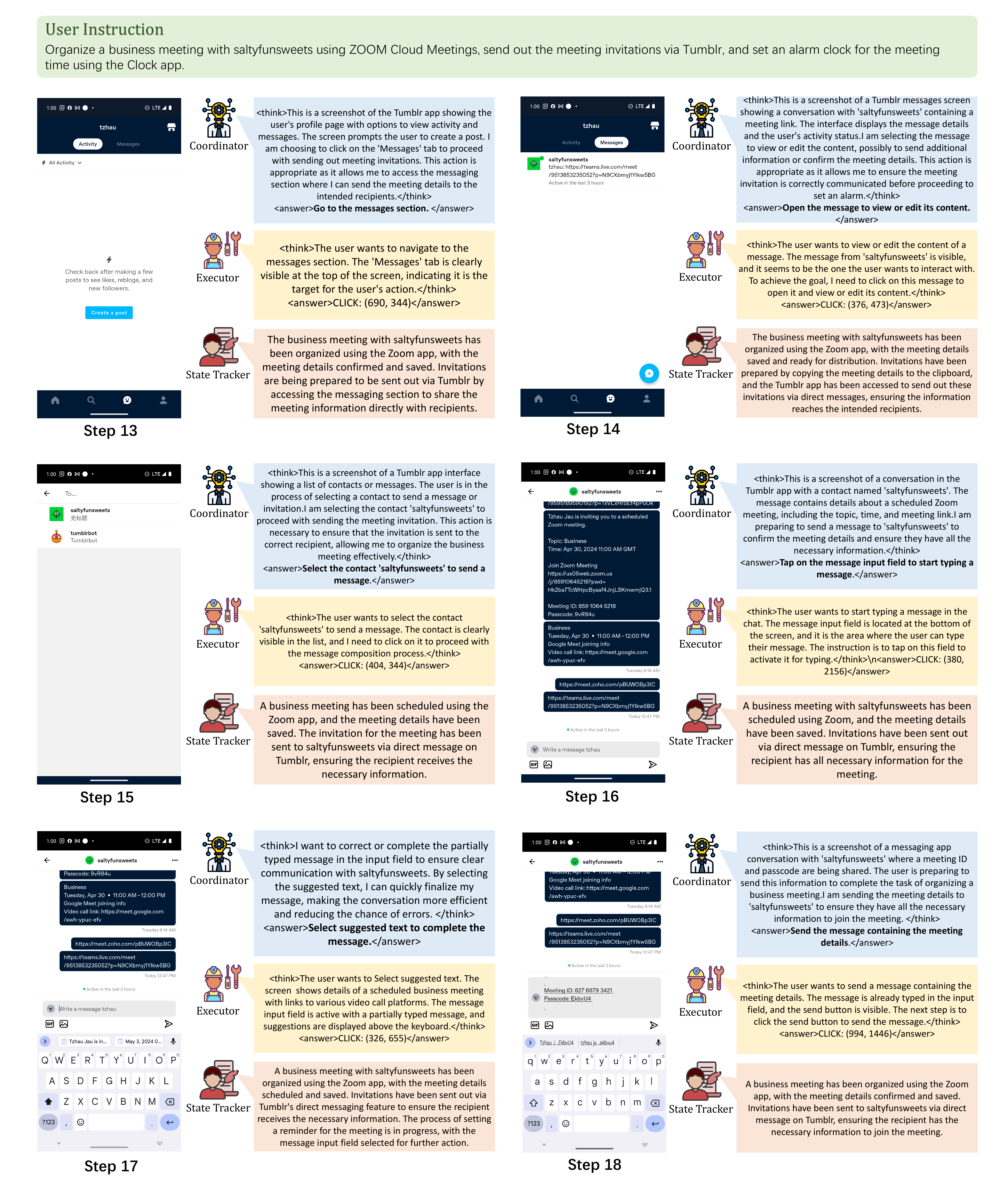}
    \caption{An example of long-horizon task (Part 3 of 5).}
    \label{fig:case3}
\end{figure}

\begin{figure}[t]
    \centering
    \includegraphics[width=\columnwidth,page=4]{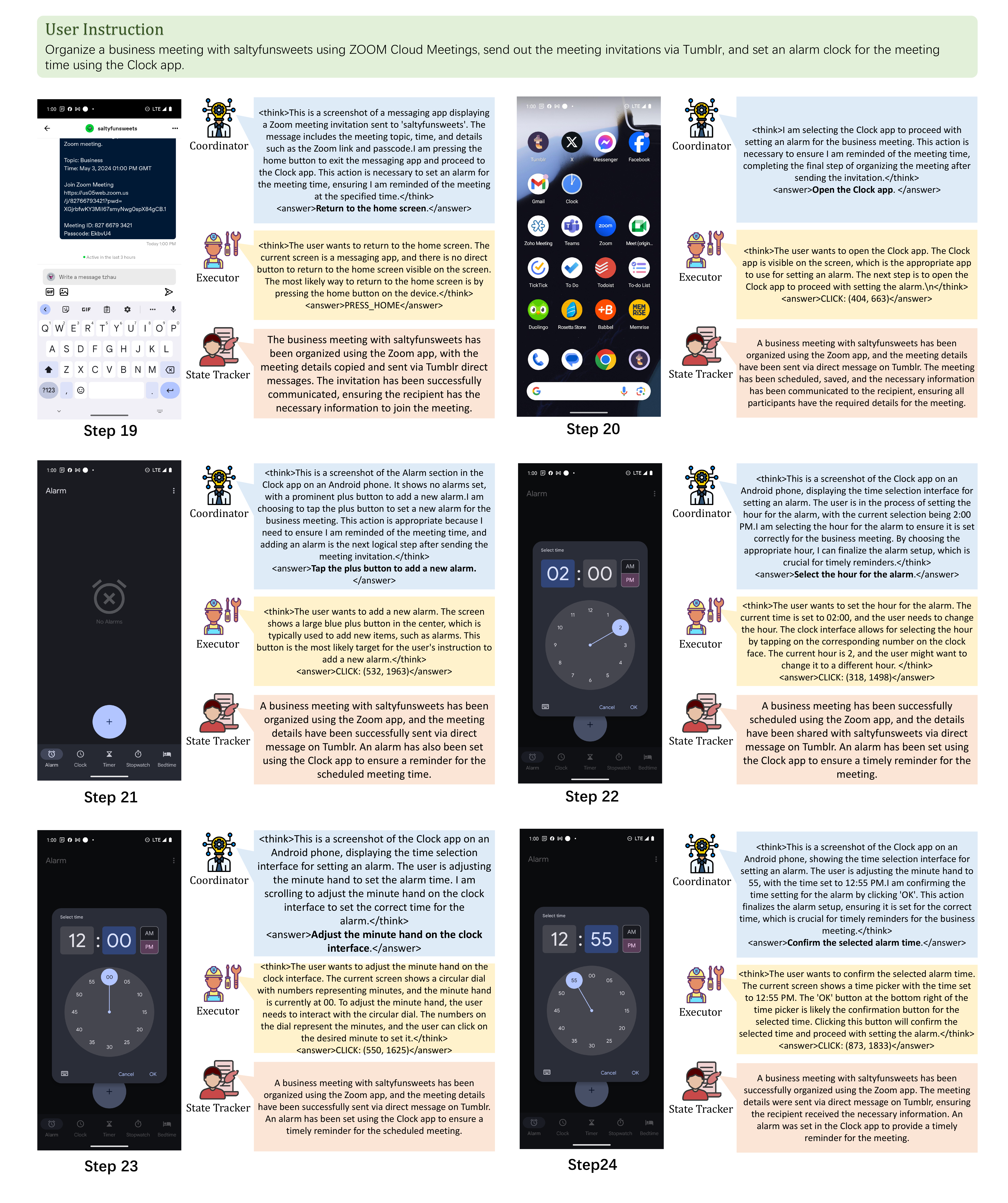}
    \caption{An example of long-horizon task (Part 4 of 5).}
    \label{fig:case4}
\end{figure}

\begin{figure}[t]
    \centering
    \includegraphics[width=\columnwidth,page=5]{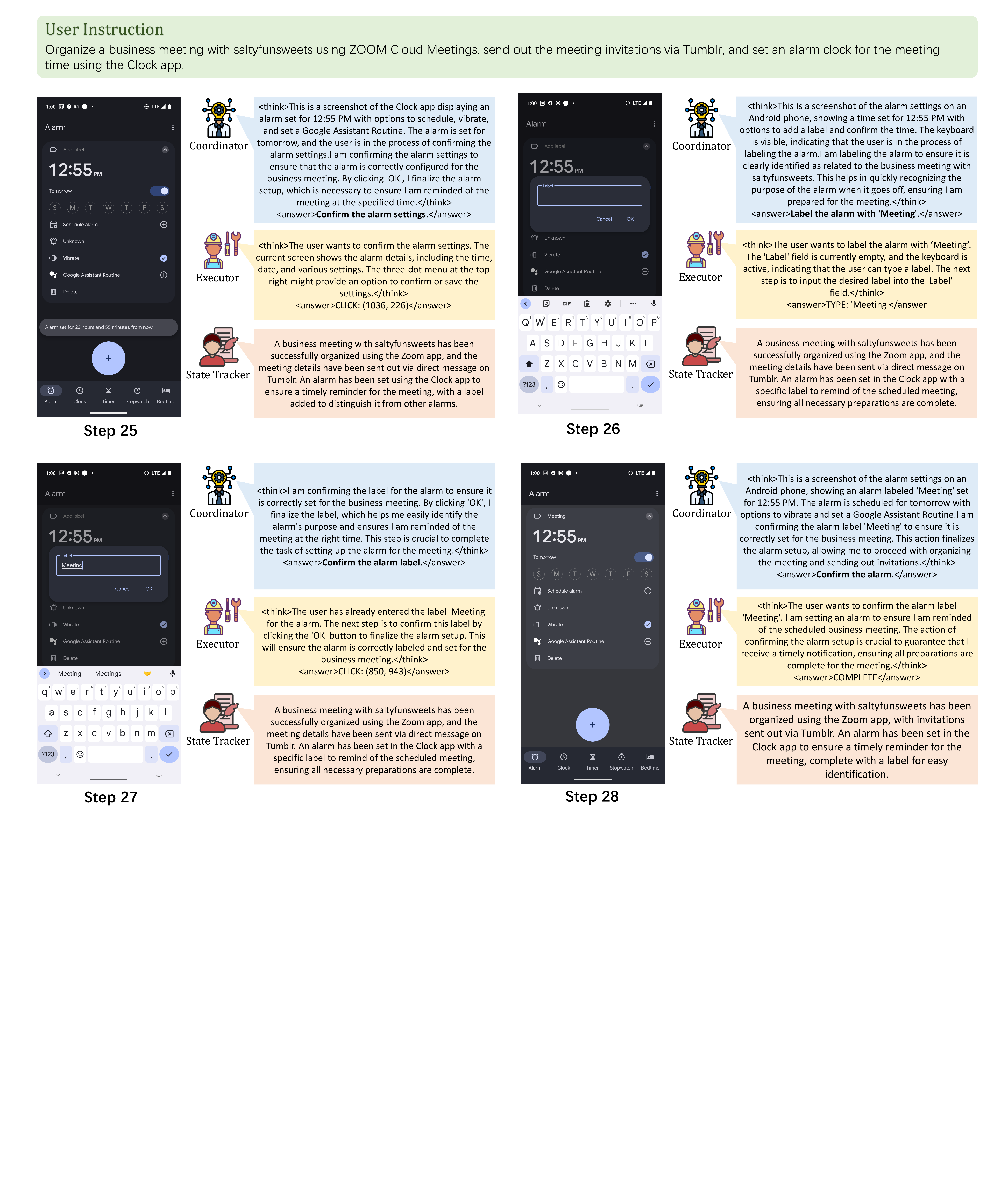}
    \caption{An example of long-horizon task (Part 5 of 5).}
    \label{fig:case5}
\end{figure}

\Cref{fig:case1}, \ref{fig:case2}, \ref{fig:case3}, \ref{fig:case4}, \ref{fig:case5} show a scenario where our CES framework solves a complete long-horizon task. 

\subsection{Overall Analysis}
The trajectory demonstrates the CES framework's ability of solving long-horizon task, driven primarily by the State Tracker’s pivotal role in bridging cross-application context gaps. For instance, after the Executor copied the meeting link in Zoom at Step 9, the agent transitioned to a completely different environment, Tumblr, in Step 12. Crucially, the State Tracker maintained a high-semantic summary: "The meeting information was copied ... to facilitate sharing", explicitly carrying this hidden state to Step 18 to guide the message sending, effectively preventing the context loss often seen in single-agent baselines during app switching. Complementing this memory retention, the Coordinator provided the necessary strategic backbone by decomposing the dense, high-level user instruction into a strictly ordered sequence: finalizing the meeting setup first, handling the information transfer second, and finally configuring the Clock in Step 28, ensuring the system maintained both logical continuity and procedural order throughout the complex workflow.

\subsection{Failure Case Analysis}
\begin{figure}[h!]
    \centering
    \includegraphics[width=\columnwidth,page=1]{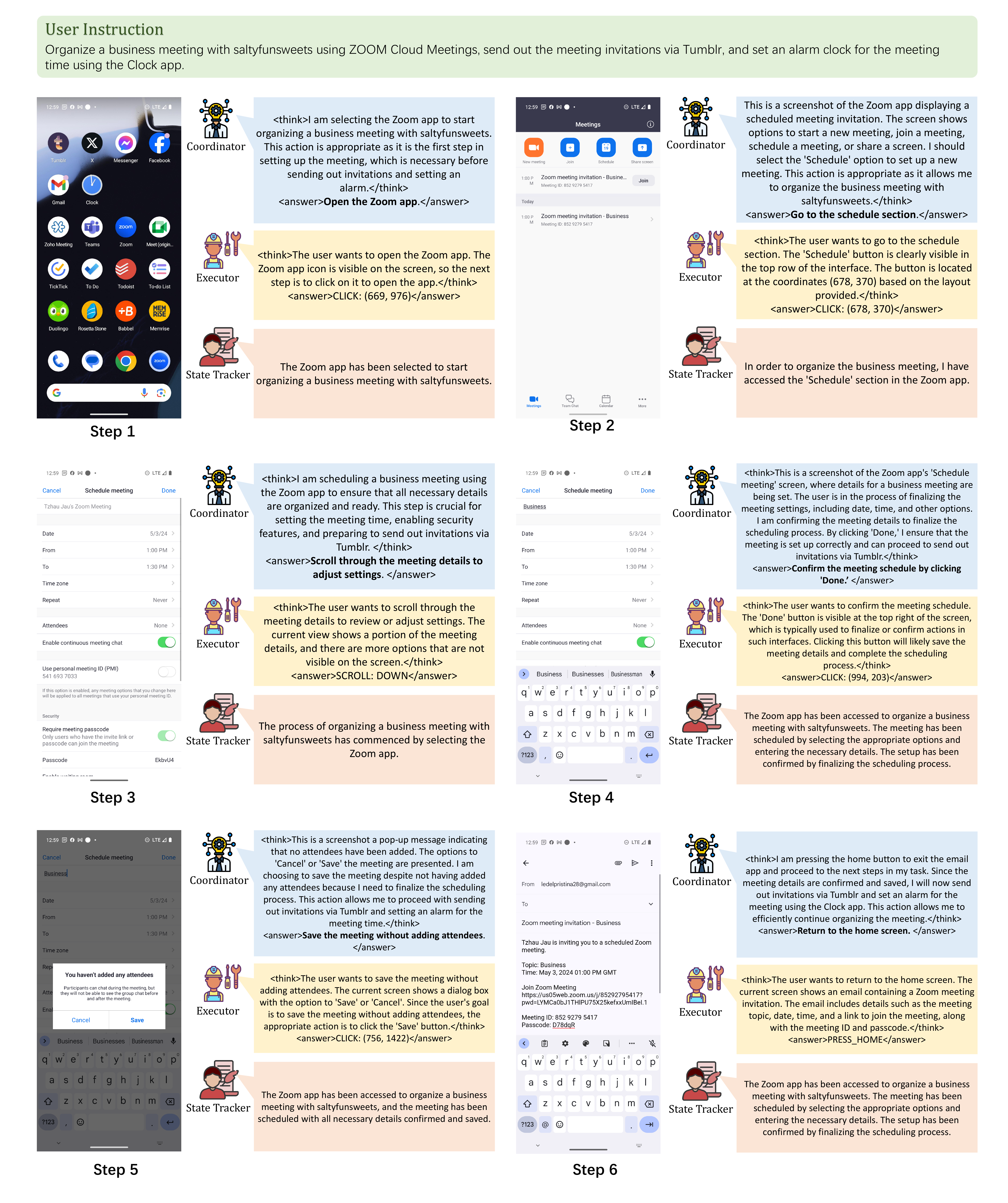}
    \caption{An example of long-horizon task (Part 1 of 5).}
    \label{fig:case1}
\end{figure}

Step 3 in \Cref{fig:case1} shows a failure case. The true action should be  \texttt{TYPE:Business}, but predicted action is \texttt{SCROLL:DOWN}. 
While the user's high-level instruction explicitly requested to ``Organize a business meeting", the Coordinator failed to translate the adjective ``business" into the specific atomic action of renaming the meeting topic. As observed in the Coordinator's output, the agent prioritized procedural configurations, such as ``setting the meeting time" and ``enabling security features", instead of noticing topic, which leads to faulty atomic instruction. 

\begin{figure}[t]
    \centering
    \includegraphics[width=\columnwidth,page=2]{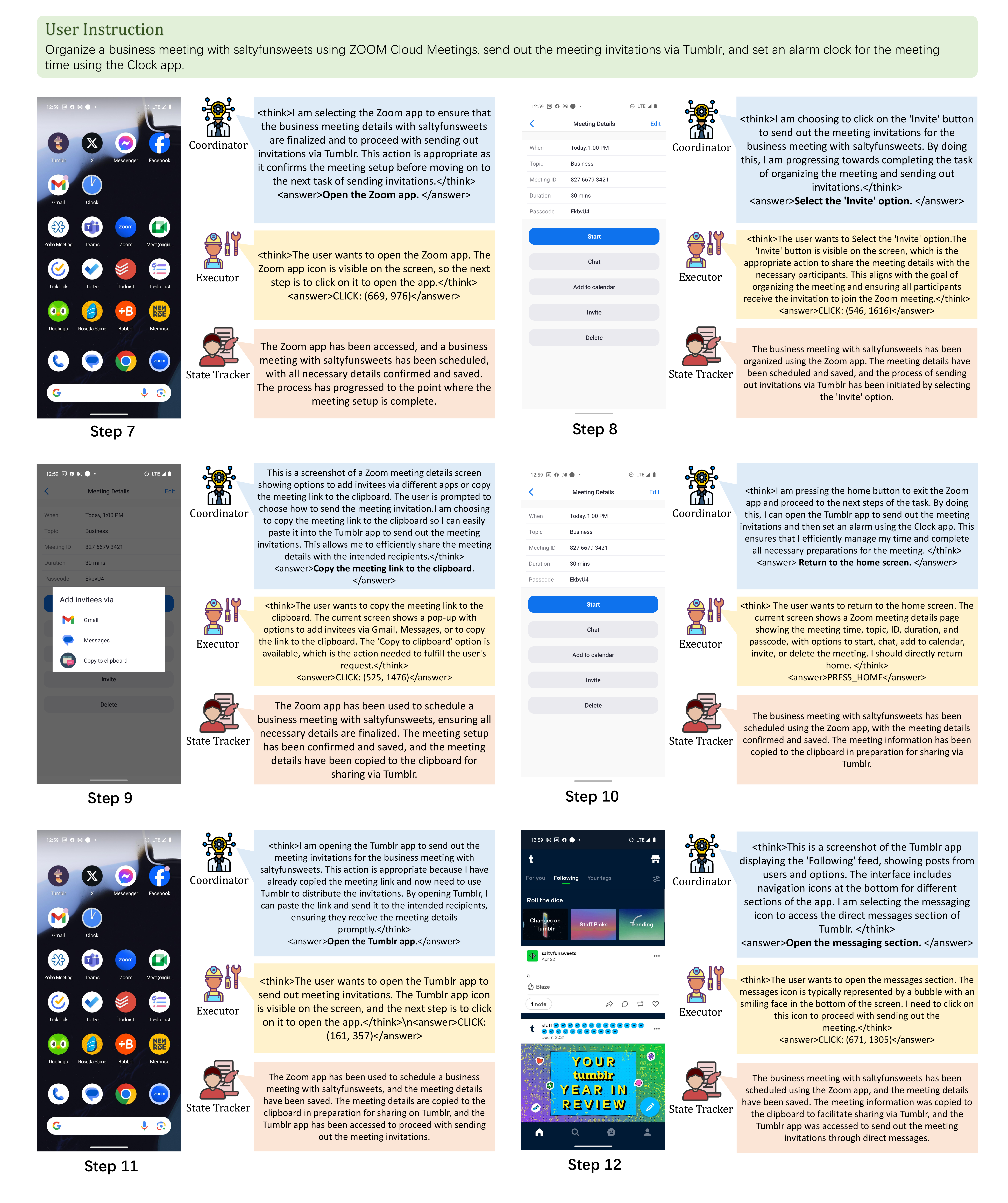}
    \caption{An example of long-horizon task (Part 2 of 5).}
    \label{fig:case2}
\end{figure}

Step 12 in \Cref{fig:case2} shows another failure case. In this step, the Coordinator correctly issued the atomic instruction "Open the messaging section". However, the Executor failed to ground this instruction to the correct pixel coordinates. A closer look at Executor's output shows that, it even correctly said ``The messages icon is typically represented by a bubble with an smiling face in the bottom of the screen", but when he answer it, he chose another bubble without smiling face in the middle of screen.

\end{document}